\documentclass[
]{ceurart}

\usepackage{listings}
\usepackage{tcolorbox}
\tcbuselibrary{skins}
\usepackage{graphicx}
\usepackage{enumitem}
\usepackage{verbatim}

\begin{document}

\copyrightyear{2025}
\copyrightclause{Copyright for this paper by its authors.
  Use permitted under Creative Commons License Attribution 4.0
  International (CC BY 4.0).}

\conference{Proceedings of the First Argument Mining and Empirical Legal Research Workshop (AMELR 2025), June 20, 2025, Chicago, United States}

\title{Generating Legal Commentaries from Case Databases via Retrieval, Clustering, and Generation}

\tnotemark[1]
\author[1]{Max Prior}[%
orcid=0009-0005-7066-996X,
email=max.prior@tum.de,
]
\address[1]{Technical University of Munich,
  Boltzmannstraße 3, 85748 Garching near Munich}
  
\author[1]{Niklas Wais}[%
orcid=0009-0008-6459-721X,
email=niklas.wais@tum.de,
]

\author[1]{Matthias Grabmair}[%
orcid=0000-0001-6586-2486,
email=matthias.grabmair@tum.de,
url=https://www.cs.cit.tum.de/lt/tum-legal-tech-working-group/,
]

\begin{abstract}
We present a fully automated pipeline that transforms large collections of court decisions into legal commentaries for statutes — without providing any handcrafted doctrinal framework. Using 4.555 decisions of the German Federal Court of Justice that cite §§ 242, 280, 812 and 823 of the German Civil Code (BGB), we extract paragraph-level chunks, summarize their reasoning, and derive keywords, which are embedded and clustered. For each cluster, an LLM generates headings and synthesizes citation-rich sections, which are then merged into coherent commentaries by four state-of-the-art LLMs. We evaluate along five dimensions -- topical relevance, heading-match, citation faithfulness, cluster distinction and logical ordering -- using both a human expert and an “LLM-judge”. Our results show that commentary-like argument mining from court decisions to generate reports that can be refreshed within minutes at minimal cost is feasible, yet they highlight limitations arising from restricted sources and the normativity of legal reasoning.
\end{abstract}


\maketitle

\section{Introduction}

Common and Civil Law systems are often contrasted by the claim that the former relies on precedent, while the latter centers on statutes. While Civil Law judges are not formally bound by case law and begin with statutes, the notion of judges as mere \textit{bouche de la loi} ("mouth of the law"; for a critical assessment of \textit{Montesquieu's} actual position, see \cite{Schonfeld_2008}) has long been proven to be a deliberate misrepresentation invented by the self-proclaimed anti-formalist movements of the early 20th century \cite{Schmidt}. Judicial decision-making necessarily involves interpretation, and prior rulings — although generally not binding — inform the consistent application of statutes. Legal commentaries systematize this practice: they collect and analyze the application of statutes by the courts as well as proposals regarding their application from papers, books, and other commentaries. In other words, \textbf{they mine arguments} from court decisions and the legal literature.

\subsection{Structure of Legal Commentaries}\label{subsec:legal commentaries}

Commentaries typically focus on one statute book, e.g., the German Civil Code (BGB), and are following its structure. In the case of a BGB commentary, there is one "chapter" per legal provision of the BGB (§ 1 to § 2385). After repeating the wording of the respective legal provision, each chapter is structured around questions arising from the application (i.e., interpretation) of the provision.
For example, when faced with a case potentially involving liability for injury under § 823 (1) BGB, a judge might question whether witnessing a traumatic event can constitute a violation of "Gesundheit" (health) as intended by the provision. Since the statutory wording itself leaves room for interpretation, the judge will seek arguments supporting or rejecting the view that psychological trauma from witnessing an event should qualify as a health violation within the meaning of § 823 (1) BGB.
This involves identifying definitions proposed by legal scholars or established by court rulings concerning the concept of "Gesundheit" (health) under § 823 (1) BGB. It also involves reviewing prior court decisions addressing similar circumstances. Regarding the question, both pro and contra arguments are collected and systematized in the § 823 BGB chapter of a commentary.

\subsection{Computer-Generated Commentaries}
Legal Commentaries are massive books written by multiple authors. Their creation is a complicated and costly process. Given the recent advances in Natural Language Processing brought about by LLMs, researchers have started to investigate the automated generation of such commentaries based on collections of court rulings.
\textit{Engel and Kruse}
present a prototype of an automated pipeline that lets GPT-4 draft a commentary on Article 8 of the German Constitution. Python code fetches 125 rulings of the German Constitutional Court, extracts passages mentioning Article 8, and feeds them -- with an elaborate prompt that entails constitutional law doctrine -- to GPT-4, which classifies, summarizes and groups every citation under typical headings. Compared to five leading hand-written commentaries, the machine-generated version cites far more decisions, provides pinpoint references and can refresh in hours at a cost of approximately \$33, yet it still misclusters some material, and misses doctrinal shifts. It was created with heavily engineered prompts that provide the doctrinal framework of German constitutional law to the model, which structures the generation into a form comparable to human-written commentaries. The authors present it as a complement, not a substitute \cite{engel2024kommentar}. They also use a similar approach with rulings from the European Court of Human Rights (ECtHR) \cite{engel2024professor}. \textit{Santosh et al.} developed the "LexGenie" model, which is not explicitly positioned as a legal commentary, but an interactive tool to generate structured reports from ECtHR rulings based on keywords provided by the user. Offline, they index at the paragraph level via Mistral-7B key-phrase embeddings and store them in a FAISS database. Online, they retrieve passages via Maximum Marginal Relevance (MMR) to the keywords and organize them using BERTopic and HDBSCAN. Afterwards, GPT-4o-mini refines headings and incrementally drafts citation-rich texts to build multi-case legal guides. Limitations include occasional misclustering and sparse cross-section links \cite{santosh2025lexgenieautomatedgenerationstructured}. Similar functionality is now being explored by systems like OpenAI’s Deep Research, which provide grounded and multi-document syntheses to support advanced academic workflows \cite{openai2025deepresearch}. Like \cite{santosh2025lexgenieautomatedgenerationstructured}, they do not rely on hard-coded legal knowledge when generating legal reports and do not follow the structure of a legal commentary. A key difference compared to \cite{santosh2025lexgenieautomatedgenerationstructured} is that they first generate a structured plan and then retrieve documents, rather than generating a structure from retrieved documents.

\subsection{Our Contribution}
We combine the approaches of \cite{santosh2025lexgenieautomatedgenerationstructured} and \cite{engel2024kommentar} by generating commentary-like reports for legal provisions (not keywords like \cite{santosh2025lexgenieautomatedgenerationstructured}) without providing doctrinal context (unlike \cite{engel2024kommentar}). Instead, we rely solely on the texts of court rulings and statutes. This generalization makes our approach easily extendable to any number of legal provisions. For demonstration purposes, we focus on the \textit{German Civil Code} (BGB), which codifies central aspects of private law and consists of more than two thousand provisions. When compared to the German Constitution or the ECHR, it is of high practical importance and particularly challenging due to the amount of court rulings that refer to it. We also compare the performance of multiple state-of-the-art models for the task at hand, introduce a sophisticated pipeline and provide quantitative as well as qualitative evaluation. Lastly, we contribute to the legal theory discussion on the benefits and limitations of computer-generated commentaries.

\section{Data and Methods}
 
\begin{figure}
    \centering
    \includegraphics[width=1\linewidth]{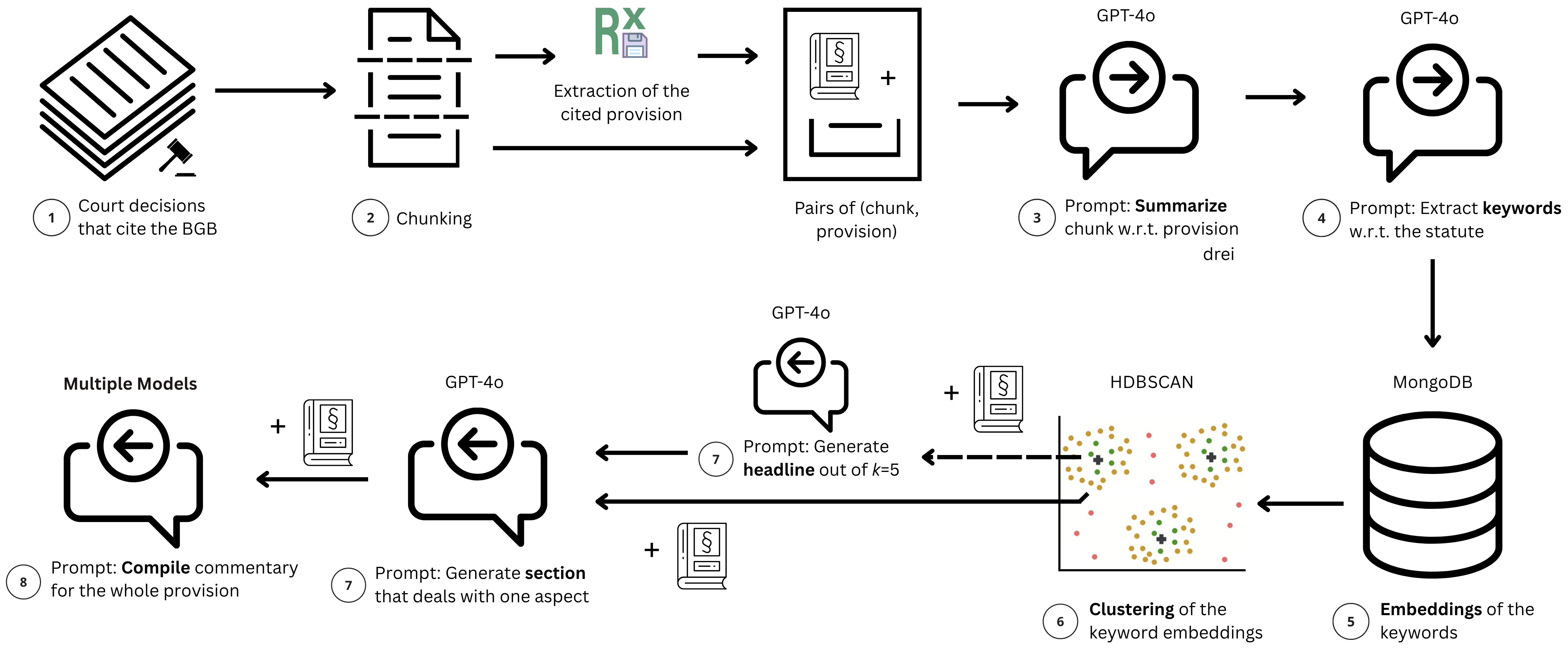}
    \caption{Our pipeline. In the clustering step (6), green circles represent the chunks used to generate headlines, while brown circles represent the broader cluster of chunks that is used to generate the section text for every headline. Red circles are regarded as noise.}
    \label{fig:pipeline}
\end{figure}

Figure \ref{fig:pipeline} illustrates the steps from downloading the court decision up to generating the final commentary.
In step 1, we select a set of practically relevant BGB provisions: § 242 \textit{Good faith}, § 280 \textit{Damages for breach of duty}, § 812 \textit{Unjust enrichment}, and § 823 \textit{Tort liability}. They are the most cited provisions in court rulings and provide some diversity: §§ 280, 812, 823 BGB form the basis for obligation-related damages and tort claims, § 242 BGB acts as a general guiding principle of good faith and fair dealing, which applies to all aspects of contract law and obligations. We rely on court rulings by the \textit{Federal Court of Justice} (BGH), Germany’s highest court for civil and criminal matters. It hears appeals from lower courts to ensure uniform interpretation of the law.   
We downloaded all 4.555 publicly available BGH decisions that cite at least one BGB provision from the German “Case Law Online” portal.\footnote{\url{https://www.rechtsprechung-im-internet.de/}} A provision is cited in approximately 11 decisions on average, but the median is only 3, i.e.\ half of all BGB provisions appear in $\leq$ 3 court decisions. The four chosen legal provisions were cited far more frequently: in 509 (§ 823), 484 (§ 280), 357 (§ 242), and 260 decisions (§ 812). In step 2, we extract the “\textit{Reasons for Decision}”  section from every court decision, split it into paragraph-level chunks, and filter out paragraphs shorter than 100 characters (they mainly contain signatures or headings). This yields the following corpus sizes: § 280 (3.191 paragraphs; 927.333 tokens), § 242 (2.872; 824.678), § 823 (2.513; 785.704), and § 812 (2.390; 663.587).
In step 3, we summarize each paragraph with GPT-4o, instructing the model to focus on how the paragraph applies the cited provision. Our pipeline automatically adds the text of the provision as context; in the case of multiple cited provisions, we provide all texts to enable the model to determine whether the chunk actually deals with the provision in question. From the summaries, we use GPT-4o to extract keywords that reflect the step of the provision's application, again automatically providing the text of the provision as context (step 4).
The keywords are embedded with \texttt{text-embedding-3-large} (step 5) and clustered per provision using HDBSCAN \cite{Campello} (step 6) -- an approach inspired by \cite{santosh2025lexgenieautomatedgenerationstructured}. The records that form the clusters have the structure: <keyword, summary, embedding>. Clustering happens in embedding space and is formed based on the semantic similarity of the keywords. Each cluster must consist of at least 20 records, and records that cannot be assigned to a cluster, shown as red dots in Figure \ref{fig:pipeline}, are considered as outliers and not further processed. Each cluster has a centroid, marked as a black cross. Orange dots depict regular cluster members, green dots highlight the five records whose keywords are closest to the centroid. These keywords are used to generate the headlines, while the summaries of all the records within a cluster are used to generate the paragraph of this section (step 7). All headline–paragraph pairs are fed to the generative models GPT-4o, GPT-4.1, GPT-4.5, and the reasoning model o3 with instructions to merge them into a well-structured commentary (step 8). For details, see Prompt 1 in the Appendix.
We applied this pipeline to generate four commentaries per model (§§ 242, 280, 812, and 823 BGB); however, the framework is generally applicable to all BGB provisions and can easily be extended to other statute books.

\section{Evaluation}
We use both LLM-as-a-judge (see Prompt 2 in the Appendix) and human evaluation to assess the overall quality of our generated commentaries (\ref{overallQuality}). In addition, we provide an in-depth (qualitative) legal analysis of the generated texts (\ref{legalAnalysis}).

\subsection{Overall Quality}\label{overallQuality}
We evaluate the overall quality based on five criteria: First, we examine topical relevance, verifying that sections align with the legal provision. Second, we check heading-matching to ensure each heading accurately reflected its section’s content. Third, we test citation faithfulness by confirming that every citation genuinely supports the statement it followed. Fourth, we inspect cluster distinction, assessing whether the sections remained clearly separated without overlap. Fifth, we review logical ordering to determine whether the document maintains a coherent, continuous narrative.
LLM-as-judge scores are provided by Gemini~2.5~Flash, which was not used for generation. Human evaluation is carried out by a member of our team with legal education in a blind process. The results of both are shown in Table \ref{tab:llm_judge}.


\renewcommand{\arraystretch}{0.9}  
\renewcommand{\arraystretch}{0.9}  
\begin{table}

§ 242 BGB  
\centering
\scalebox{0.80}{%
\begin{tabular}{lccccc}
  \toprule
  \textbf{Model} & \textbf{Topical Relevance} & \textbf{Heading-Match} &
  \textbf{Citation-Faithfulness} & \textbf{Cluster-Distinction} & \textbf{Logical Ordering}\\
  \midrule
  GPT-4.1         & 3 | 5 & 4 | 5 & 3 | 4 & 4 | 4 & 3 | 4 \\
  GPT-4.5-preview & 5 | 5 & 5 | 5 & 4 | 4 & 4 | 5 & 4 | 4 \\
  GPT-4o          & 3 | 4 & 3 | 5 & 4 | 4 & 3 | 3 & 3 | 3 \\
  o3              & 4 | 4 & 5 | 5 & 3 | 2 & 5 | 4 & 5 | 4 \\
  \bottomrule
\end{tabular}}

§ 280 BGB  
\scalebox{0.80}{%
\begin{tabular}{lccccc}
  \toprule
  \textbf{Model} & \textbf{Topical Relevance} & \textbf{Heading-Match} &
  \textbf{Citation-Faithfulness} & \textbf{Cluster-Distinction} & \textbf{Logical Ordering}\\
  \midrule
  GPT-4.1         & 3 | 5 & 4 | 4 & 4 | 4 & 3 | 5 & 3 | 5 \\
  GPT-4.5-preview & 3 | 5 & 5 | 4 & 3 | 4 & 5 | 5 & 5 | 5 \\
  GPT-4o          & 2 | 5 & 4 | 5 & 4 | 2 & 3 | 5 & 3 | 5 \\
  o3              & 3 | 5 & 5 | 4 & 3 | 5 & 4 | 4 & 5 | 5 \\
  \bottomrule
\end{tabular}}

§ 812 BGB  
\scalebox{0.8}{%
\begin{tabular}{lccccc}
  \toprule
  \textbf{Model} & \textbf{Topical Relevance} & \textbf{Heading-Match} &
  \textbf{Citation-Faithfulness} & \textbf{Cluster-Distinction} & \textbf{Logical Ordering}\\
  \midrule
  GPT-4.1         & 4 | 4 & 5 | 3 & 4 | 2 & 2 | 3 & 3 | 2 \\
  GPT-4.5-preview & 5 | 4 & 5 | 5 & 4 | 2 & 4 | 3 & 4 | 4 \\
  GPT-4o          & 3 | 3 & 4 | 2 & 3 | 1 & 2 | 1 & 3 | 2 \\
  o3              & 3 | 4 & 5 | 5 & 3 | 3 & 5 | 5 & 5 | 5 \\
  \bottomrule
\end{tabular}}

§ 823 BGB  
\scalebox{0.8}{%
\begin{tabular}{lccccc}
  \toprule
  \textbf{Model} & \textbf{Topical Relevance} & \textbf{Heading-Match} &
  \textbf{Citation-Faithfulness} & \textbf{Cluster-Distinction} & \textbf{Logical Ordering}\\
  \midrule
  GPT-4.1         & 4 | 4 & 4 | 5 & 4 | 3 & 4 | 5 & 3 | 3 \\
  GPT-4.5-preview & 5 | 5 & 5 | 4 & 4 | 3 & 5 | 4 & 4 | 4 \\
  GPT-4o          & 3 | 2 & 3 | 5 & 3 | 4 & 2 | 2 & 2 | 2 \\
  o3              & 3 | 4 & 5 | 5 & 4 | 3 & 5 | 5 & 4 | 5 \\
  \bottomrule
\end{tabular}}

Average score across legal provisions  
\scalebox{0.80}{%
\begin{tabular}{lccccc}
  \toprule
  \textbf{Model} & \textbf{Topical Relevance} & \textbf{Heading-Match} &
  \textbf{Citation-Faithfulness} & \textbf{Cluster-Distinction} & \textbf{Logical Ordering}\\
  \midrule
  GPT-4.1         & 3.50 | 4.50 & 4.25 | 4.25 & \textbf{3.75} | \textbf{3.25} & 3.25 | 4.25 & 3.00 | 3.50 \\
  GPT-4.5-preview & \textbf{4.50} | \textbf{4.75} & \textbf{5.00} | 4.50 & \textbf{3.75} | \textbf{3.25} & 4.50 | 4.25 & 4.25 | 4.25 \\
  GPT-4o          & 2.75 | 3.50 & 3.50 | 4.25 & 3.50 | 2.75 & 2.50 | 2.75 & 2.75 | 3.00 \\
  o3              & 3.25 | 4.25 & \textbf{5.00} | \textbf{4.75} & 3.25 | \textbf{3.25} & \textbf{4.75} | \textbf{4.50} & \textbf{4.75} | \textbf{4.75} \\
  \bottomrule
\end{tabular}}

\caption{Evaluation scores for \S~242\,BGB, \S~280\,BGB, \S~812\,BGB, \S~823\,BGB and their overall average (1-5, the higher the better).  
For every model–criterion pair, human-annotated scores appear on the left, LLM-based scores on the right (human | LLM).}
\label{tab:llm_judge}
\end{table}
GPT-4.5-preview leads with an average human evaluation score of 4.4 across all criteria and provisions, closely followed by o3 (4.2), while GPT-4o trails in every category with an average score of 3. The human judge rates GPT-4.5-preview highest in topical relevance and award both GPT-4.5-preview and o3 scores ranging from 4 to 5 for heading-match, cluster-distinction, and logical order. However, all models plateau around 3.5 for citation faithfulness. Gemini 2.5 Flash consistently scores commentaries higher by approximately half a point in topical relevance, cluster-distinction, and ordering. The closest alignment between model and human evaluations appears in citation faithfulness, where the automated model scores slightly lower. Divergence between human and automated scores is most pronounced for § 812 BGB, with high human scores contrasting sharply with low LLM ratings. §§ 242 and 280 BGB show closer human-model agreement. The evaluation highlights three findings: First, GPT-4.5-preview and GPT-4o generate high-quality commentaries but have shortcomings in citation faithfulness. Second, the reasoning model o3 trails behind in topical relevance, but shows significant improvements in cluster distinction and logical ordering. Here, GPT-4o occupies the middle ground between GPT-4o and o3, pointing towards a sweet spot in a potential trade-off between the quality of the structure and the content. Third, the substantial gap between human and LLM scores casts doubt on the reliability of model-based evaluations. We have not measured inter-rater agreement, which we leave for future work.

\subsection{Legal Analysis}
An in-depth qualitative comparison of the generated texts with traditional legal commentaries reveals similarities and peculiarities.\footnote{The generated reports are available at \url{https://github.com/amelr250501/icail}}\label{legalAnalysis} First, the structure of the reports for §§~280, 823, 812 BGB on the one hand, and §~242 BGB on the other hand, diverge -- like in traditional commentaries. This is because the former are bases for claims, while the latter is a general provision with broad use cases. Provisions that function as the basis for claims provide the structure for their examination, which is reflected in traditional commentaries. All models mimic this quite accurately for §~280 BGB, with occasional hickups in the ordering. § 823 BGB and § 812 BGB are more challenging in this regard, because they combine multiple claims, each with diverging structures. GPT-4.1 and GPT-4o struggle with their separation, while GPT-4.5-preview manages to account for them, although not perfectly. This is where o3 shines. It presents two different structures for §~823 BGB and comes up with a surprising solution for §~812 BGB: After presenting key points, it switches to forming case groups. Breaking away from the dogmatic analysis appears reasonable in an area of the law where even the BGH acknowledges that certain cases "cannot be assessed in a general way, but must rather be determined on a case-by-case basis, taking into account the specific circumstances of the respective case" \cite{BGH_2006_III_ZR_62_05}. The problem with o3, however, is that it places too much emphasis on structure; the result sometimes resembles a collection of claim structures rather than a commentary. GPT-4.5-preview does a better job at incorporating more substance in the form of dogmatic explanations and example cases. Legal practitioners will find the results to be closer to a traditional commentary. Nevertheless, even GPT-4.5-preview remains on the "legal surface". In particular, all generated texts lack discussions of specific problems of interpretation, in which different opinions (and arguments supporting them) are weighed against each other. This is what commentaries are frequently consulted for.

\section{Discussion and Limitations}
When compared to traditional, hand-written legal commentaries, their new generated counterparts display some promising features. They are able to handle massive amounts of sources. Furthermore, their creation is way cheaper, which allows developers to offer them at a lower price point; we publish our results at no cost. In contrast, hand-written commentaries are expensive. Physical copies have to be re-bought with new editions. While subscription-based online versions exist, their actualization still requires manual labor and is therefore slow. In contrast, our automated commentary can be re-generated with every new decision within minutes. Unlike \cite{engel2024kommentar}, its creation does not require any specific legal knowledge and can therefore be easily extended to all areas of law. In comparison to \cite{santosh2025lexgenieautomatedgenerationstructured}, it nevertheless manages to recognize the dogmatic structures found in conventional commentaries. Does the future of legal commentaries therefore lie in the machine-made processing of court rulings? In light of our findings, we argue that LLMs can be a great help for collecting as well as evaluating sources and the general structuring of legal commentaries. If one takes into account the ever-growing body of statutes and court rulings, their usage might even become necessary. The proposed multi-step setup with chunking, abstract summarizations, keyword generations, clustering, and text generation has led to impressive results in our experiments. However, this can only form a starting point and is not without limitations. As such, we identify the practical problem of limited sources (\ref{limitedSources}) and a fundamental problem of value judgments (\ref{valueJudgements}).

\subsection{Limited Sources}\label{limitedSources}
Like [1] and [2], we mine arguments from court decisions (whose judge authors presumably consulted legal commentaries, which in turn evaluated the legal literature). Directly mining the legal literature would provide an unfiltered, more diverse set of arguments. Their absence becomes obvious in the lack of any reported conflicts of opinion in the commentaries generated by us.
The underlying problem is a practical one caused by the limited access to commercial databases in Germany.
\textit{Engel} and \textit{Kruse} as well as \textit{Santosh et al.} are less affected by it, because both the court rulings from the Federal Constitutional Court of Germany and the European Court of Human Rights are binding -- a rare case in Civil Law. Even if a ruling is considered to be law-creating (e.g., by \textit{Kelsen's} ``Pure Theory of Law''), the generated norm is individual; it does not become a general norm unless the legal system explicitly says otherwise. This is the case for both courts (§ 31[1] BVerfGG and Art. 46[1] EMRK). Conversely, court rulings of the BGH or arguments put forward in them carry, in theory, the same weight as arguments put forward by any lower court or legal scholar. While practitioners will nevertheless be guided by the ``case law'' of the BGH, arguments for contestation put forward in the literature are helpful where a -- possible -- deviating outcome is sought. Also, the literature deals with cases that have not yet been decided by a court. Well-founded concepts proposed by legal scholars can form the starting point of the decision-making process if the respective cases emerge in practice.

\subsection{Value Judgments}\label{valueJudgements}
This points to a conceptual problem that arises from the fact that, to be helpful in practice, commentaries are actually \textit{expected} to be opinionated. \textit{Christoph Engel} and \textit{Johannes Kruse} argue that using LLMs with predefined inputs renders the creation of commentaries more objective \cite{engel2024kommentar}. Since our commentaries present their reports as mere summaries (i.e., "detached observations" \cite{Fletcher}), one can go further and call this approach "scientific" in the strict sense of \textit{Hans Kelsen}.
Computer-generated commentaries like ours adhere to the ideal he has set for unauthentic interpretation provided by legal commentaries -- they come close to what he regarded as "scientific commentaries" \cite{kelsen1951law} that refrain from “committed arguments” \cite{Fletcher}.
However, traditional legal commentaries are opinionated in their reasoning \textit{because court rulings and pleadings themselves are opinionated} in their reasoning. Commentaries serve as a practical tool; whether, like \textit{Kelsen}, one wishes to deny them scientific character on that basis is a separate question. Practitioners seek from them a well-founded interpretation, fully aware that alternative interpretations may exist. Switching the style of the generated output from "detached observations" to "committed arguments", however, raises a serious question: When mining arguments from multiple sources, whose value judgments form the basis of the proposed "right" solution? In this context, the use of the supposedly intelligent and objective LLM could even revive the long overcome belief that the one "right" solution can be found by technical means.

\section{Conclusions}
This paper demonstrates that when used in carefully designed pipelines, LLMs can create doctrine-oriented legal commentaries at the speed with which new court decisions appear. Starting from 4.555 rulings of the German Federal Court of Justice, our system is capable of creating reports for §§ 242, 280, 812 and 823 of the German Civil Code without handcrafted doctrinal scaffolding -- and can be extended to any number of provisions. In a blind review, our expert rates the best variant (GPT-4.5-preview) at 4.4/5 across relevance, structure and logical order. Citation grounding remains the chief weakness. Compared with traditional hand-written commentaries, the machine-generated versions can cover more decisions, can be regenerated within minutes and cost orders of magnitude less. Yet, they do not contain the "commited arguments" legal practitioners will typically expect.

\section*{Acknowledgements}
This work was carried out within the project “Generatives Sprachmodell der Justiz (GSJ)”, a joint initiative of the Ministry of Justice of North Rhine‑Westphalia (Ministerium der Justiz des Landes Nordrhein‑Westfalen) and the Bavarian State Ministry of Justice (Bayerisches Staatsministerium der Justiz), with the scientific partners Technical University of Munich (Technische Universität München) and University of Cologne (Universität zu Köln). The project is financed through the Digitalisierungsinitiative des Bundes für die Justiz.

\section*{Declaration on Generative AI}
  During the preparation of this work, the author(s) used ChatGPT in order to: Grammar and spelling check, Paraphrase and reword. After using this tool/service, the author(s) reviewed and edited the content as needed and take(s) full responsibility for the publication’s content.
  \newline

\bibliography{sample-ceur}

\section*{Appendix}


\begin{center}

\begin{tcolorbox}[
    width=\dimexpr\linewidth+2cm\relax,
    enlarge left by=-0.5cm,
    colframe=black,
    colback=white,
    boxrule=0.5pt,
    arc=0pt,
    outer arc=0pt,
    enhanced,
    attach boxed title to bottom center={yshift=-2mm},
    boxed title style={colframe=white, colback=white, boxrule=0pt},
    fonttitle=\small\bfseries,
    title={Prompt (English translation)}
    ]
\textbf{You are a German attorney.}

Consider the preceding legislative text ("normative text") only to the extent necessary to correctly anchor definitions, structure, and the telos of the norm; do not quote it extensively or summarize it literally.

Revise the following draft linguistically, restructure it logically, number and name the headings consistently, so that each section logically builds upon the previous one. Connect transitions clearly and avoid redundancy. Avoid generic headings (e.g., \textit{"Concept," "Practical Case"}). Do not include concluding summaries.

The commentary should reflect the abstract structure of the application of the norm but should also include concrete examples if they appear in the draft. Such examples should only relate to selected aspects of the norm. Do not invent examples; only use those included in the draft.

Write in an objective, formal style. This commentary is intended for trained legal practitioners, not for students. The output may be longer than the input. Avoid bullet points; instead, formulate their content in full sentences.

IMPORTANT: Each occurrence of ObjectId('…') corresponds to a reference and must remain as such in the final text if retained.

Return \textbf{only the commentary}.
\end{tcolorbox}

\begin{tcolorbox}[
    width=\dimexpr\linewidth+2cm\relax,
    enlarge left by=-0.5cm,
    colframe=black,
    colback=white,
    boxrule=0.5pt,
    arc=0pt,
    outer arc=0pt,
    enhanced,
    attach boxed title to bottom center={yshift=-2mm},
    boxed title style={colframe=white, colback=white, boxrule=0pt},
    fonttitle=\small\bfseries,
    title={Prompt (Deutsch)}]
\textbf{Du bist ein deutscher Rechtsanwalt.}
Berücksichtige den vorangestellten Gesetzestext („Normtext“) lediglich in angemessenem Umfang, um Definitionen, Systematik und Telos der Norm korrekt zu verankern; zitiere ihn nicht ausführlich und fasse ihn nicht wörtlich zusammen.

Überarbeite den folgenden Entwurf sprachlich, strukturiere ihn logisch um, nummeriere und benenne die Überschriften konsistent, sodass jeder Abschnitt sinnvoll auf den vorangehenden aufbaut. Verbinde Übergänge, vermeide Redundanz. Vermeide generische Überschriften (z.B. \textit{"Begriff", "Praxisfall"}). Verzichte auf abschließende Zusammenfassungen.

Der Kommentar soll die abstrakte Struktur der Anwendung der Norm widerspiegeln, aber auch konkrete Beispiele beinhalten, wenn diese im Entwurf vorkommen. Solche Beispiele sollen sich nur auf ausgewählte Aspekte der Norm beziehen. Erfinde keine Beispiele, sondern greife nur in dem Entwurf enthaltene Beispiele auf.

Schreibe in einem sachlichen, formalen Stil. Es handelt sich um einen Kommentar für ausgebildete Rechtsanwender, nicht etwa für Studierende. Der Output darf länger sein als der Input. Verzichte auf Stichpunkte, formuliere deren Inhalt stattdessen aus.

WICHTIG: Jede Stelle ObjectId('…') entspricht einer Referenz und muss als solche im finalen Text erhalten bleiben, falls der Text übernommen wird.

Gib \textbf{nur den Kommentar} zurück.
\end{tcolorbox}

\small\textbf{Prompt 1: Generation of the final commentary, German original and translation}
\end{center}

\begin{tcolorbox}[
    width=\dimexpr\linewidth+1cm\relax,
    enlarge left by=-0.5cm,
    colframe=black,
    colback=white,
    boxrule=0.5pt,
    arc=0pt,
    outer arc=0pt,
    enhanced,
    attach boxed title to bottom center={yshift=-2mm},
    boxed title style={colframe=white, colback=white, boxrule=0pt},
    fonttitle=\small\bfseries,
    title={Prompt (English Translation)}
    ]
\textbf{Evaluation Criteria}

Critically evaluate the text of a German legal commentary and assign a score from \textbf{1 (barely satisfactory)} to \textbf{5 (very well fulfilled)} for each of the following criteria:

\begin{enumerate}
    \item \textbf{Topical Relevance:} Do the headings cover all undefined terms of the underlying norm?
    \item \textbf{Heading-Match:} Does each paragraph fully meet the content promised by its heading?
    \item \textbf{Citation-Faithfulness:} Do the cited references genuinely support the statements (no hallucinations)? Are all referenced documents locatable?
    \item \textbf{Cluster-Distinction:} Is the content clearly distinct with minimal or no overlap with other sections within the text (clear thematic demarcation)?
    \item \textbf{Logical Ordering:} Does the placement of each section logically fit into the overall structure (coherent thread, comprehensible sequence)?
\end{enumerate}

Return only the result in JSON format—without any additional text.
\end{tcolorbox}

\begin{tcolorbox}[
    width=\dimexpr\linewidth+1cm\relax,
    enlarge left by=-0.5cm,
    colframe=black,
    colback=white,
    boxrule=0.5pt,
    arc=0pt,
    outer arc=0pt,
    enhanced,
    attach boxed title to bottom center={yshift=-2mm},
    boxed title style={colframe=white, colback=white, boxrule=0pt},
    fonttitle=\small\bfseries,
    title={Prompt (Deutsch)}
    ]
\textbf{Bewertungsrichtlinien}

Bewerte den Text eines deutschen juristischen Kommentars \textbf{kritisch} und vergebe für jedes der folgenden Kriterien einen Wert von \textbf{1 (befriedigt kaum)} bis \textbf{5 (sehr gut erfüllt)}:

\begin{enumerate}
    \item \textbf{Topical Relevance:} Decken die Überschriften alle unbestimmten Begriffe der zugrunde liegenden Norm ab?
    \item \textbf{Heading-Match:} Entspricht jeder Absatz inhaltlich vollständig dem Versprechen der Überschrift?
    \item \textbf{Citation-Faithfulness:} Stützen die angegebenen Fundstellen die Aussagen tatsächlich (keine Halluzinationen)? Werden alle referenzierten Dokumente gefunden?
    \item \textbf{Cluster-Distinction:} Deckt sich der Inhalt nicht oder nur minimal mit anderen Abschnitten innerhalb des Texts (klare thematische Abgrenzung)?
    \item \textbf{Logical Ordering:} Passt die Position jedes Abschnitts in die Gesamtstruktur (roter Faden, nachvollziehbare Reihenfolge)?
\end{enumerate}

Gib ausschließlich das Ergebnis im JSON-Format zurück – ohne weiteren Text.
\end{tcolorbox}
\small\textbf{Prompt 2: Evaluation of the commentary. German original and translation}









\end{document}